\UseRawInputEncoding
\documentclass[sigconf]{acmart}
\usepackage[ruled,vlined]{algorithm2e}
\usepackage{multirow}

\newcommand{\best}[1]{{\textbf{#1}}} 
\newcommand{\second}[1]{{\underline{#1}}} 

\newcommand{\method}[1]{{#1}}
\newcommand{\setting}{{UPFL}}
\newcommand{\algo}{\method{FedTTA}}
\newcommand{\algoprox}{\method{FedTTA-Prox}}
\newcommand{\algopp}{\method{FedTTA++}}
\newcommand{\heteroalgo}{\method{HeteroFedTTA}}
\AtBeginDocument{%
  }

\setcopyright{acmcopyright}
\copyrightyear{2018}
\acmYear{2018}
\acmDOI{XXXXXXX.XXXXXXX}

\acmConference[Conference acronym 'XX]{Make sure to enter the correct
  conference title from your rights confirmation emai}{June 03--05,
  2018}{Woodstock, NY}
\acmPrice{15.00}
\acmISBN{978-1-4503-XXXX-X/18/06}

\begin{document}

\title{
UPFL: Unsupervised 
Personalized Federated Learning 
towards 
New Clients}

\author{Tiandi Ye}
\affiliation{%
 \institution{East China Normal University,}
 \city{}
 \state{Shanghai}
 \country{China}
}
\email{52205903002@stu.ecnu.edu.cn}

\author{Cen Chen}
\affiliation{%
 \institution{East China Normal University,}
 \city{}
 \state{Shanghai}
 \country{China}
}
\email{cenchen@dase.ecnu.edu.cn}

\author{Yinggui Wang}
\affiliation{%
 \institution{Ant Group}
 \city{}
 \state{Beijing}
 \country{China}
}
\email{wyinggui@gmail.com}

\author{Xiang Li}
\affiliation{%
 \institution{East China Normal University,}
 \city{}
 \state{Shanghai}
 \country{China}
}
\email{xiangli@dase.ecnu.edu.cn}

\author{Ming Gao}
\affiliation{%
 \institution{East China Normal University,}
 \city{}
 \state{Shanghai}
 \country{China}
}
\email{mgao@dase.ecnu.edu.cn}

\renewcommand{\shortauthors}{Tiandi Ye, Cen Chen, Yinggui Wang, Xiang Li and Ming Gao.}

\begin{abstract}
Personalized federated learning has gained significant attention as a promising approach to address the challenge of data heterogeneity. In this paper, we address a relatively unexplored problem in federated learning. 
When a federated model has been trained and deployed, and an unlabeled new client joins, providing a personalized model for the new client becomes a highly challenging task. To address this challenge, we extend the adaptive risk minimization technique into the unsupervised personalized federated learning setting and propose our method, \algo. We further improve \algo~with two simple yet effective optimization strategies: enhancing the training of the adaptation model with proxy regularization and early-stopping the adaptation through entropy. Moreover, we propose a knowledge distillation loss specifically designed for \algo~to address the device heterogeneity. Extensive experiments on five datasets against eleven baselines demonstrate the effectiveness of our proposed \algo~and its variants. The code is available at: https://github.com/anonymous-federated-learning/code. 
\end{abstract}



\received{20 February 2007}
\received[revised]{12 March 2009}
\received[accepted]{5 June 2009}

\maketitle

\section{Introduction}\label{sec:introduction}
Federated learning (FL) is a popular machine learning paradigm, in which multiple clients collaboratively train machine learning models without compromising their data privacy~\cite{kairouz2021advances}. 
During each FL communication round, the server synchronizes the latest global model with {a subset of selected clients. These selected clients} then optimize the model on their local data and send back the updated model parameters to the server. 
The server then aggregates the received model parameters 
and updates the global model. 

{Data heterogeneity often poses a significant challenge to FL}, where the data distribution among clients is non-independent and non-identically distributed (non-IID). {This non-IID nature of the data may significantly impact the model performance.} 
{To address this challenge, personalized federated learning has emerged as a promising approach}, as it allows for a personalized model for \textit{each participating client}~\cite{arivazhagan2019federated,fallah2020personalized,shamsian2021personalized}. 
{Most of the works focus on improving the performance of the global model by tailoring it to the specific data distribution of each client, while }
few works have considered how to provide personalized models for clients that are not available during training time, i.e., \textit{new clients}. 
\method{Per-FedAvg}~\cite{fallah2020personalized} and \method{pFedHN}~\cite{shamsian2021personalized} have made some explorations, which personalize the model for new clients by fine-tuning on their \textit{labeled datasets}. 
However, during inference time, new clients often lack labeled data, which motivates us to investigate the unsupervised settings.

In this paper, we consider a practical yet more challenging task {for heterogeneous federated learning}: 
\textit{
\underline{U}nsupervised 
\underline{P}ersonalized 
\underline{F}ederated 
\underline{L}earning towards new clients  (\setting)}. 
Specifically, 
after a federated model has been trained and deployed, 
new clients may wish to join and utilize the deployed model for inferring their local \textit{unlabeled} datasets. 
The data distribution of these new clients might significantly differ from that of the training clients. 
Providing personalized models for these new clients with \textit{only unlabeled datasets} is a highly challenging task.
To the best of our knowledge, 
the only work that addresses the same task as ours is \method{ODPFL-HN}~\cite{amosy2021inference} , which extends \method{pFedHN} to the 
\textsc{UPFL} setting.
Specifically, \method{ODPFL-HN} trains an encoder network to learn a representation for each client based on its unlabeled data, 
and then feeds the client representation to a hypernetwork that generates a personalized model for that client. 
{However,} \method{ODPFL-HN} involves 
{additionally transmitting the client representation, which could potentially put the clients' privacy at risk.
And it would generate} the same model structure for all clients, which limits its applicability in heterogeneous FL settings. 

{To effectively address the above-mentioned \setting~task, we leverage the idea of adaptive risk minimization~\cite{zhang2021adaptive} into federated learning and propose our approach \algo. }
Specifically, each client $c_{i}$ meta-learns a base prediction model $f(\cdot|\psi^{i})$ 
and an adaptation model $g(\cdot|\phi^{i})$, which takes in the outputs of $f(\cdot|\psi^{i})$ on unlabeled data points and produces personalized parameters $\tilde{\psi^{i}}$. 
The server aggregates the base prediction models and adaptation models from the clients to update the server base prediction model $\psi^{s}$ and adaptation model $\phi^{s}$. When a new client with unlabeled data joins, it downloads the server models $\psi^{s}$ and $\phi^{s}$, and adapts $\psi^{s}$ to a personalized one $\tilde{\psi^{s}}$ under the supervision of $\phi^{s}$. 

{We further introduce an improved version of \algo, denoted as \algopp, which leverages} two simple yet effective optimization strategies: 
(1) enhancing the adaptation model's personalization capability with regularization on the base prediction model and 
(2) early stopping the adaptation during testing based on the entropy of the prediction model's outputs. 
{First, the base prediction model in FL can easily overfit due to local data scarcity. Moreover,}
a highly personalized base prediction model could already perform very well on the local dataset, 
{making it difficult} for the adaptation model to learn the ability to customize the base prediction model. To address this, we introduce regularization on the outputs of the base prediction model to enhance the training of the adaptation model's personalization capability. 
{Second, we observe that only \textit{one-step} update may result in an under-explored personalization model, while excessively updating the base prediction model during adaptation can lead to sub-optimal solutions.}
{To address this, we propose to utilize the entropy of the personalized model's predictions as an unsupervised metric for early stopping of the adaptation process, as personalized models generally exhibit low entropy in their predictions. }
Specifically, we stop the adaptation when the entropy of the personalized model's predictions does not decrease within a certain steps. 

Moreover, device heterogeneity is also prevalent in federated learning, with clients exhibiting varying computing and storage capabilities, thereby resulting in model structures and sizes that differ significantly.
To address device heterogeneity, we develop a heterogeneous version of \algo,~called \heteroalgo, based on a classic heterogeneous federated learning framework \method{FedMD}~\cite{li2019fedmd} that enables knowledge transfer between heterogeneous clients through knowledge distillation. 
Based on \algo, a straightforward approach is to distill knowledge for student's base prediction model and adaptation model in an end-to-end manner, with the aim of approximating the predictions of the student's personalized model to those of the teacher.
{However in \algo,} the knowledge from teacher model {may arise} from two sources: the general prediction capability of the base prediction model and the adaptation capability of the adaptation model in transforming a base prediction model into a personalized one. 
Thus, we suggest {a novel knowledge distillation loss that enforces the student's} base prediction model and adaptation model to learn from the corresponding models of the teacher, respectively. 

We summarize our contributions as follows: 
\begin{itemize}
    \item {We address a practical and yet rarely studied task in federated learning, namely \setting, where the primary challenge} is to personalize models for new unlabeled clients. To address this challenge, we propose \algo. 
    \item We further propose \algopp, an enhanced version of \algo, which leverages two simple 
     optimization strategies to achieve remarkable performance improvements. 
    \item Considering the heterogeneous model setting and the unique architecture of \algo, we suggest a novel knowledge distillation loss, which is empirically shown to be effective. 
    \item We conduct extensive experiments {on five benchmark datasets against eleven baselines} to evaluate the effectiveness of our proposed methods, and the results demonstrate their efficacy in addressing the \setting$\;$task.
\end{itemize}

\section{Related Work}
\noindent \textbf{Personalized federated learning (PFL)} 
customizes each client with a personalized model based on its local data~\cite{arivazhagan2019federated,fallah2020personalized,shamsian2021personalized}. 
PFL has gained significant advancements recently. 
For example, 
\method{FedPer}~\cite{arivazhagan2019federated} learns a universal representation layer across clients and personalizes a local model with a unique classification head. 
\method{pFedHN}~\cite{shamsian2021personalized} leverages hyper-network to output a personalized model for each client, 
and \method{Per-FedAvg}~\cite{fallah2020personalized} coordinates clients to learn a global model, from which each client can quickly adapt to a personalized model through a one-step update. 
Most existing PFL methods focus solely on improving the performance of training clients, 
neglecting the crucial aspect of generalization to new clients. 
Although \method{pFedHN} and \method{Per-FedAvg} have made some attempts to address this, 
both of them rely on fine-tuning on new clients' labeled datasets, which does not apply to our task.

\noindent \textbf{Generalized federated learning (GFL)} is proposed to enhance the performance of new clients 
by learning invariant mechanisms among training clients 
and expecting them to generalize well to new clients~\cite{tenison2022gradient,nguyen2022fedsr,liu2021feddg,zhang2021federated,caldarola2022improving,qu2022generalized}. 
For example, 
\method{FedGMA}~\cite{tenison2022gradient} presents a gradient-masked averaging approach to better capture the invariant mechanism across heterogeneous clients. 
\method{FedSR}~\cite{nguyen2022fedsr} implicitly aligns the marginal and conditional distribution of the representation with two regularizers. 
A similar idea has also been exploited in \method{FedADG}~\cite{zhang2021federated}, which adaptively learns a dynamic reference distribution to accommodate all source domains. 
GFL methods output a single model and directly apply it to test clients, 
while \algo~ is an unsupervised PFL method 
that adapts the base prediction model to a personalized model 
based on hints from the new client's unlabeled dataset. 
To the best of our knowledge, 
\method{ODPFL-HN}~\cite{amosy2021inference} is the only work that addresses the same problem as ours. 
\method{ODPFL-HN} trains 
a client encoder network 
and a hypernetwork, 
where the client encoder network encodes an unlabeled client to a representation and the hypernetwork generates a personalized model based on that representation. 

\noindent \textbf{Test-Time Adaptation (TTA)} 
refers to fine-tuning a model during inference time to adapt to specific test data. 
TTA was first introduced by Test-Time Training (\method{TTT})~\cite{sun2020test}, 
which additionally learns a self-supervised auxiliary task (rotation prediction) during training 
and fine-tunes the feature encoder based on this task during testing. 
Since then, test-time adaptation methods 
have been proposed one after another. 
For example, \method{TTT++}~\cite{liu2021ttt++} builds upon \method{TTT} 
and introduces a test-time feature alignment strategy. 
\method{TENT}~\cite{wang2020tent} fine-tunes the model at inference time 
to optimize the model confidence 
by minimizing the entropy of its predictions. 
However, there has been limited exploration of 
the intersection of personalized federated learning 
and test-time adaptation. 
The recently proposed work, FedTHE~\cite{jiang2022test}, studies the problem that clients' local test sets evolve during deployment and proposes to robustify federated learning models by adaptive ensembling of the global generic and local personalized classifier of a two-head federated learning model. 
This approach does not apply {to our setting, 
as new clients lack} personalized classifiers.

\noindent \textbf{Knowledge Distillation (KD)}
has been receiving increasing attention from the federated learning community, 
as it enables the aggregation of knowledge from heterogeneous models~\cite{gao2022survey}. 
Existing heterogeneous federated learning methods 
typically aggregate the logits of clients 
rather than their model parameters on the server. 
Each client then updates the local model 
to approximate the global average logits. 
For instance, 
\method{FedMD}~\cite{li2019fedmd} performs ensemble distillation on a public dataset, 
while \method{KT-pFL}~\cite{zhang2021parameterized} improves on \method{FedMD} by updating the local model with the personalized logits of each client through a parameterized knowledge coefficient matrix. 
However, our focus in this paper is not on the heterogeneous federated learning framework. 
Instead, our contribution is proposing a more effective knowledge distillation loss, 
which can better facilitate the distillation of the models in \algo.

\section{Methodology}
In this section, 
we will provide a detailed description 
of the learning setting of \setting~towards new clients. 
We will then introduce our proposed method \algo~and propose an enhanced version \algopp~, 
which leverages two simple optimization strategies: enhancing the training of the adaptation model by regularizing the outputs of the base prediction model and early stopping the adaptation process during testing through entropy. 
We will also give a heterogeneous version of \algo based on \method{FedMD}, denoted as \heteroalgo. 
We suggest a novel knowledge distillation loss for \heteroalgo, which facilitates the application of \algo~in heterogeneous federated learning setting. 

\subsection{Problem Formulation}
In this work, we focus on the scenario 
where training is performed on $N$ clients, 
denoted by $\{c_{1}, \cdots, c_{N}\}$, 
each with a private dataset $\mathcal{D}_i = \{ (x_{j}^{(i)}, y_{j}^{(i)}) \}_{j=1}^{m_i}$ of size $m_i$, 
and {new clients may} join in at the inference stage. 
Our objective is to learn an unsupervised adaptation strategy from the training clients, 
which can provide a personalized model for a new client 
based on its unlabeled dataset $\mathcal{D}_{new} = \{ x_{j}^{(i)} \}_{j=1}^{m_{new}}$. 
We refer to this learning setting as the \textit{
\underline{U}nsupervised \underline{P}ersonalized \underline{F}ederated \underline{L}earning (\setting)} 
\footnote{In this work, we primarily focus on the multi-class classification problem, but much of our analysis can be extended directly to regression and other problems. }
. 

\subsection{Algorithm \algo}
\begin{figure*}
    \centering
    \includegraphics[width=0.9\textwidth,trim=135 100 125 105,clip]{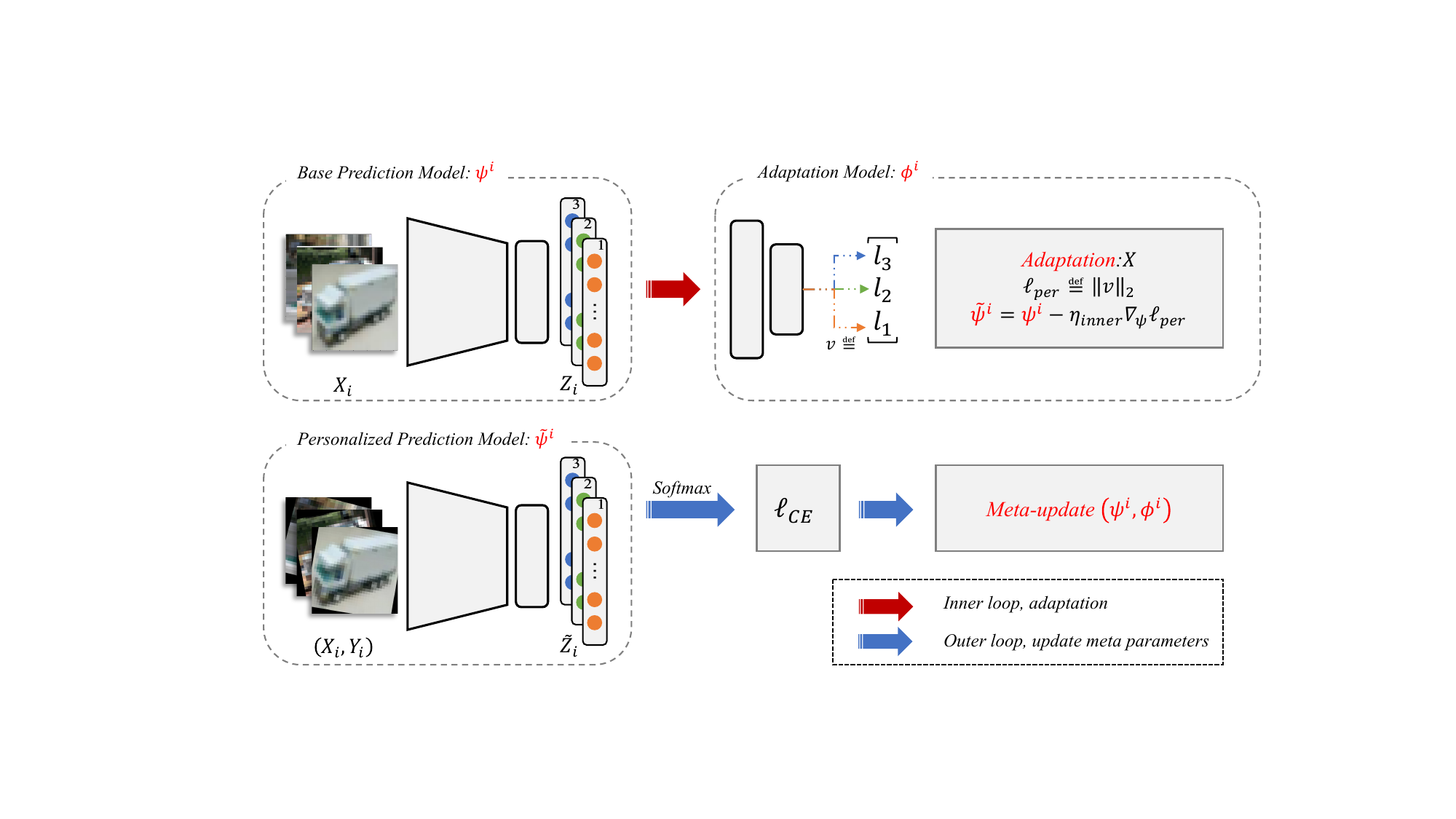}
    \caption{
    The figure depicts the local training process of client $c_i$ in \algo. 
    The inner loop outputs a personalized prediction model, while the outer loop evaluates this model and meta-updates both the base prediction model and adaptation model accordingly.}
    \label{fig:framework}
\end{figure*}

The proposed framework is illustrated in Fig.~\ref{fig:framework}. 
It can be seen from the figure that 
each client's local models 
consist of two components: 
base prediction model $f(\cdot;\psi)$ and adaptation model $g(\cdot;\phi)$. 
The base prediction model $f$ is the task network for classification, 
which takes a batch of samples $X:=\{x_1, \cdots, x_{B}\}$ 
and outputs the corresponding logits $Z:=\{z_1, \cdots, z_{B}\}$
~\footnote{For the convenience of description, we regard all the data of the client as a batch here.}: 
\begin{equation}
Z = f(X, \psi)
\end{equation}
The adaptation model $g$ is responsible for 
customizing the base prediction model 
to a personalized prediction model. 
$g$ takes the outputs of the base prediction model 
and outputs a scalar for each sample 
that measures 
how \textit{poorly} current base prediction model $f$ 
performs on corresponding unlabeled data. 
Following~\cite{zhang2021adaptive}, 
we use the $\ell_{2}$-norm of these scalars 
across the batch 
as the personalization loss of $f$ on the client. 
Adaptation is performed by updating $f(\cdot;\psi)$ towards minimizing the personalization loss. 
Specifically, 
\begin{equation}
\ell_{per} = \Vert g(Z; \phi) \Vert_{2}
\label{eq:personalization-loss}
\end{equation}
Obviously, the output of the adaptation model 
is independent of samples order. 
Regardless of how the samples are sorted, 
the adaptation model's evaluation 
of the base prediction model's performance 
is consistent. 

\subsubsection{Training Phase}
During training, each client has access to its labeled dataset 
$\mathcal{D}_i := \{ (x_{j}^{(i)}, y_{j}^{(i)}) \}_{j=1}^{m_i}$. 
We denote $X_{i} := \{ x_{j}^{(i)} \}_{j=1}^{m_i}$ and $Y_i := \{ y_{j}^{(i)} \}_{j=1}^{m_i}$. 
We simulate the testing environment 
to meta-train the base prediction model 
and the adaptation model. 
Specifically, we first adapt the base prediction model towards a personalized model: 
\begin{equation}
Z_{i} = f({X}_i; \psi^{i})
\end{equation}
\begin{equation}
\ell_{per} = \Vert g(Z_{i}; \phi^{i}) \Vert_2
\end{equation}
\begin{equation}
\tilde{\psi}^{i} \leftarrow \psi^{i} - \eta_{inner} \nabla_{\psi} \ell_{per}, 
\end{equation}
where $\eta_{inner}$ is the learning rate of the prediction model in the inner loop. 
Then, 
we optimize the parameters $\psi^{i}$ and $\phi^{i}$ to minimize the loss of the personalized model $\tilde{\psi}^{i}$ on its local dataset: 
\begin{equation}
\tilde{Z_{i}} = f({X}_i; \tilde{\psi}^{i})
\end{equation}
\begin{equation}
\mathcal{L} = \ell_{CE}(\tilde{Z_{i}}; Y_{i})
\label{eq:meta-loss}
\end{equation}
\begin{equation}
\psi^{i} \leftarrow \psi^{i} - \eta_{outer} \nabla_{\psi}\mathcal{L}
\end{equation}
\begin{equation}
\phi^{i} \leftarrow \phi^{i} - \eta_{adapt} \nabla_{\phi}\mathcal{L},
\end{equation}
where $\eta_{outer}$ is the learning rate of the prediction model in the outer loop 
and $\eta_{adapt}$ denotes the learning rate of the adaptation model. 

\subsubsection{Testing Phase}
Once a new client $c_{new}$ joins 
after the federated model has been deployed, 
it downloads the models from the server 
and adapts the base prediction model 
under the supervision of the adaptation model, 
and then makes predictions 
based on the personalized prediction model:
\begin{equation}
Z_{new} = f(\mathcal{D}_{new}; \psi^{new})
\end{equation}

\begin{equation}
\psi^{new} \leftarrow \psi^{new} - \eta_{inner} \nabla_{\psi} \Vert g(Z_{new}; \phi^{new}) \Vert_2
\end{equation}

\begin{equation}
\tilde{Z}_{new} = f(\mathcal{D}_{new}; \psi^{new})
\end{equation}

\subsection{Improved \algo~(\algopp)} 
\subsubsection{Enhance Adaptation Model with Regularization}
In federated learning, 
the client data is often limited, 
which can easily lead to overfitting 
of the base prediction model. 
A highly personalized base prediction model 
can already perform very well on the client's local data, 
making it difficult for the adaptation model to learn how to customize a general prediction model to a personalized prediction model. 
In addition, during testing, the base prediction model often gives general prediction results. 
Therefore, we expect the adaptation model to focus more on the general prediction model. 
Overfitting of the base prediction model can lead to poor generalization of the adaptation model. 
To mitigate the issue, 
during local training, 
we constrain the outputs of the local base prediction model 
to approach that of the server prediction model $\dot{Z}:=f(X_i;\psi^{s})$, 
which enables the adaptation model 
to be effectively trained to adapt a general prediction model into a personalized one. 
We call such a variant of \algo~as \algoprox. 
With the constraint, the loss function in Eq.~\ref{eq:meta-loss} becomes: 
\begin{equation}
\mathcal{L} = \ell_{CE}(\tilde{Z_{i}}; Y_{i})+\mu \ell_{KL}(\sigma(Z_{i})||\sigma(\dot{Z_{i}}))
\end{equation}
, where $\ell_{KL}$ and $\sigma(\cdot)$ denote the KL divergence and softmax function, respectively, while $\mu$ serves as a balancing coefficient. 

Different from \method{FedProx}~\cite{li2020federated}, 
which limits the local updates 
directly in parameter space for mitigating client variances, 
we regularize the base prediction model's outputs (logits) to be general (less personalized) 
and expect the adaptation model can adjust them to be personalized ones.

\subsubsection{Early Stop Adaptation through Entropy} 
\begin{figure}[h!]
    \centering{}
    \includegraphics[width=1.0\linewidth]{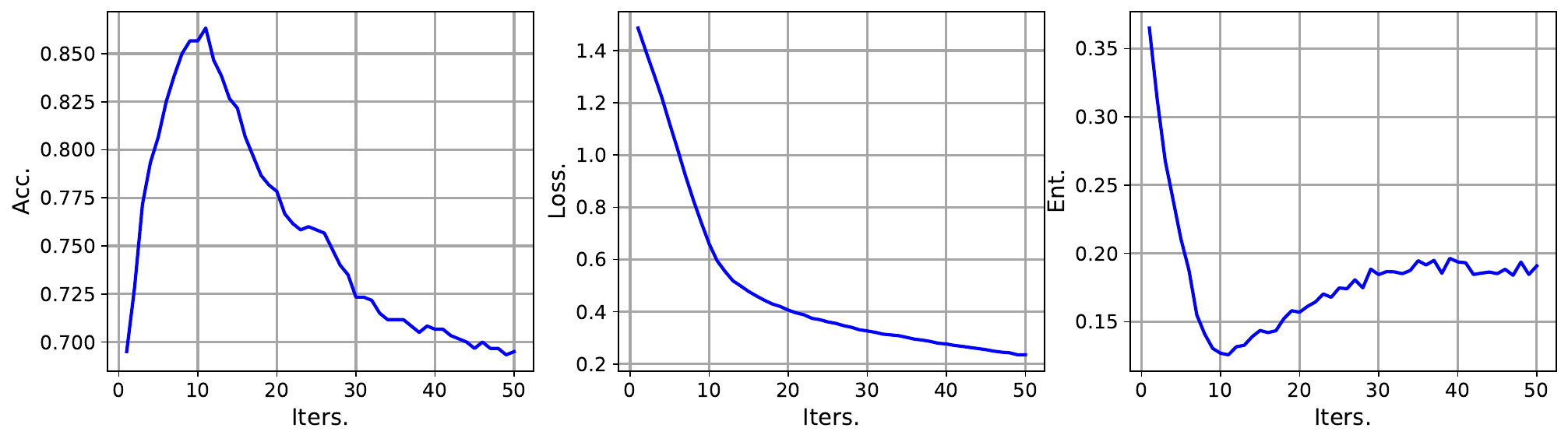}
    \caption{
    An example of mismatch between $\ell_{per}$ and the performance of the personalized prediction model. 
    The figures show 
    during adaptation a client's change curves of the personalized prediction model's accuracy (left), 
    $\ell_{per}$ (middle), 
    and the entropy of the personalized model's predictions (right). 
    While additional fine-tuning can improve the personalized model in the initial stages, excessive fine-tuning can lead to a suboptimal personalized model. 
    To prevent this, we suggest early stopping the adaptation process when the entropy reaches its lowest point (at the 12-th iteration).
    }
    \label{fig:mismatch}
\end{figure}

\algo~outputs a base prediction model and an adaptation model, 
aiming to personalize the base prediction model quickly for a new client 
through \textit{a one-step update} under the adaptation model's supervision. 
However, a \textit{one-step update} may not be sufficient, 
and \textit{multi-step} fine-tuning can often yield a better personalized model. 
We run \algo~on CIFAR-10 dataset, 
and during testing, 
each client updates the base prediction model for 50 iterations. 
Interestingly, we have empirically observed a slight mismatch 
between $\ell_{per}$ (defined in Eq.~\ref{eq:personalization-loss}) 
and the performance of the personalized model as fine-tuning progresses. 
The example in Fig.~\ref{fig:mismatch} shows that additional fine-tuning 
can improve the personalized model in the first 11 steps, 
but excessive fine-tuning can lead to a suboptimal personalized model. 
Obviously the number of fine-tuning steps 
is a crucial parameter 
that could vary among clients, 
and it is unreasonable to apply a universal hyperparameter to all clients. 

To address this, 
we suggest an adaptive early stopping strategy 
based on the \textit{entropy} 
of the personalized model's predictions. 
In general, a better personalized model will produce a more confident prediction with low entropy. 
Thus, we calculate the entropy of the personalized model's predictions during fine-tuning 
and terminate fine-tuning when the entropy does not decrease within a certain steps. 
For example, we should stop the adaptation at the 12-th iteration for the client illustrated in Fig~\ref{fig:mismatch}. 

\begin{algorithm}[h!]
\SetAlgoLined
\caption{FedTTA++}
\label{alg:algorithm}
\textbf{Input}: initial parameters of base prediction model $\psi^{0}$ and adaptation model $\phi^{0}$, 
total communication rounds $R$, 
local iterations $\tau$, 
local minibatch size $B$, 
number of participating clients per round $M$,
and maximum number of iterations for fine-tuning during testing $E$ \\
\textbf{Output}: $\psi_{R}$ and $\phi_{R}$ \\
\textbf{[Training]} \\
\For{{each round} $r$ : 0 {to} $R-1$} {
    Select a set of clients $\mathcal{S}_{r}$ of size $M$ \\
    \For{\texttt{each client} $c_{i} \in \mathcal{S}_{r}$} {
        // Local Training \\
        $\{\psi^{i}, \phi^{i}\} \leftarrow \{\psi_{r}, \phi_{r}\}$ \\
        $\mathcal{B}_{i} \leftarrow$ Split local dataset $\mathcal{D}_i$ into batches of size $B$ \\
        \For{{each iteration} $t$ : 1 {to} $\tau$} {
            Sample a batch $(X_{t}, Y_{t}) \sim \mathcal{B}_{i}$ \\
            $\dot{Z_{t}} = f(X_t; \psi_r)$ \\
            $Z_{t} = f(X_t; \psi^{i})$ \\
            $\ell_{per} = \Vert g(Z_{t}; \phi^{i}) \Vert$ \\
            $\tilde{\psi}^{i} \leftarrow \psi^{i} - \eta_{inner} \nabla_{\psi} \ell_{per}$ \\
            $\tilde{Z_{t}} = f(X_t; \tilde{\psi}^{i})$ \\
            $\mathcal{L} = \ell_{CE}(\tilde{Z_{t}}; Y_{t})+\mu \ell_{KL}(\sigma(Z_{t})||\sigma(\dot{Z_{t}}))$ \\
            $\psi^{i} \leftarrow \psi^{i} - \eta_{base} \nabla_{\psi}\mathcal{L}$ \\
            $\phi^{i} \leftarrow \phi^{i} - \eta_{adapt} \nabla_{\phi}\mathcal{L}$ \\
        }
        $\{\psi_{r+1}^{i}, \phi_{r+1}^{i}\} \leftarrow \{\psi^{i}, \phi^{i}\}$ \\
        Client $c_{i}$ sends $\psi_{r+1}^{i}$ and $\phi_{r+1}^{i}$ back to server \\
    }
    // Server Aggregation \\
    $\psi_{r+1} \leftarrow \frac{1}{M}\sum_{c_{i} \in \mathcal{S}_{r}} \psi_{r+1}^{i}$ \\
    $\phi_{r+1} \leftarrow \frac{1}{M}\sum_{c_{i} \in \mathcal{S}_{r}} \phi_{r+1}^{i}$ \\
}
\textbf{[Test of client $c_{new}$]} \\
$\{\psi^{new}, \phi^{new}\} \leftarrow \{\psi_{R}, \phi_{R}\}$ \\
\For{\texttt{each epoch} $e$: 1 \texttt{to} $E$} {
    $Z_{new} = f(\mathcal{D}_{new}; \psi^{new})$ \\
    $\psi^{new} \leftarrow \psi^{new} - \eta_{inner} \nabla_{\psi}g(Z_{new}; \phi^{new})$ \\
    $\tilde{Z}_{new} = f(\mathcal{D}_{new}; \psi^{new})$ \\
} 
Select the model whose predictions have the least entropy, i.e., $\mathcal{H}(\sigma(\tilde{Z}_{new}))$ 
\end{algorithm}

\subsection{Heterogeneous \algo~(\heteroalgo)}
For heterogeneous federated learning, 
we propose \heteroalgo, 
which allows clients to have diverse model structures. 
\heteroalgo~ \\ builds upon a classic heterogeneous federated learning framework called \method{FedMD}~\cite{li2019fedmd}, 
which facilitates knowledge transfer between heterogeneous clients through knowledge distillation on a public dataset. 
Our contribution lies in proposing a \textit{more efficient} knowledge distillation loss function for \algo~and \algopp, 
which significantly enhances the transfer of knowledge in generating personalized models. 

Before presenting the details of our proposed loss function, 
we provide a brief overview of how \heteroalgo~runs under the framework of \method{FedMD}. 
\method{FedMD} assumes that all clients have access to a public \textit{unlabeled} dataset 
$\mathcal{D}_p:=\{ x_{j}^{(p)} \}_{j=1}^{m_p}$ 
that can be collected from public sources or generated using generative methods~\cite{Goodfellow2014GenerativeAN,zhu2021data}. 
During training, the server sends each client the ensemble knowledge about $\mathcal{D}_p$, 
i.e., the average of the logits 
produced by clients' personalized prediction models on $\mathcal{D}_p$. 
Each client first digests the ensemble knowledge via knowledge distillation 
and then meta-trains local models on its labeled dataset. 
Upon completion of local training, 
each client obtains the personalized prediction model 
by adapting its base prediction model 
under the adaptation model's supervision 
and then sends the logits to the server 
made by the personalized prediction model on $\mathcal{D}_p$. 
The server aggregates received logits and updates the ensemble knowledge about $\mathcal{D}_p$. 
When a new client joins, 
it can customize the architecture 
of its base prediction model 
and adaptation model 
and then distills the ensemble knowledge about the public dataset to its local models. 

Efficient knowledge distillation is crucial for \heteroalgo. 
Each client locally trains a pair of models, 
including a base prediction model and an adaptation model. 
For ease of description, 
we denote the base prediction model, personalized prediction model and adaptation model of the teacher and student as 
$\{\psi^t, \tilde{\psi}^t, \phi^t\}$ and 
$\{\psi^s, \tilde{\psi}^s, \phi^s\}$. 
A straightforward knowledge distillation approach 
for \algo~and \algopp~
is to meta-train ${\psi^s, \phi^s}$ \textit{end-to-end} 
such that the outputs of $\tilde{\psi}^s$ 
approximate that of $\tilde{\psi}^t$ as closely as possible. 
However, 
we believe that 
the teacher's knowledge comes from two sources: 
the general prediction ability of the base prediction model 
and the adaptation model's personalization capacity. 
The \textit{end-to-end} knowledge distillation might result in insufficient training of $\phi^s$, 
as $\psi^{s}$ might lazily learn from $\tilde{\psi}^t$. 

To address this, 
we propose that 
the student should 
enforce its base prediction model and adaptation model learn from the corresponding models of the teacher, respectively. 
Specifically, 
$\psi^{s}$ should mimic the outputs of $\psi^{t}$, 
while $\phi^{s}$ should learn the personalization capability of $\phi^{t}$. 
Under the constraint of 
minimizing the KL-divergence between the outputs of $\psi^{s}$ and $\psi^{t}$, 
$\phi^{s}$ can learn the personalization ability of $\phi^{t}$ 
by minimizing the KL-divergence between the outputs of $\tilde{\psi}^{s}$ and $\tilde{\psi}^{t}$. 
We formulate the knowledge distillation loss function as: 
\begin{equation}
\label{eq:kd_loss}
\begin{split}
\mathcal{L}_{KD}(\psi^{s},\phi^{s}) = 
\sum_{x\in \mathcal{D}_{p}} \ell_{KL} (\sigma(f(x|\tilde{\psi}^{s})), \sigma(f(x|\tilde{\psi}^{t}))) \\ + 
\lambda \ell_{KL} (\sigma(f(x|\psi^{s})), \sigma(f(x|\psi^{t})))
\end{split}
\end{equation}
, where $\ell_{KL}$ denotes the KL-divergence and $\sigma(\cdot)$ denotes the softmax function. 
We provide a schematic diagram of the knowledge distillation loss in the appendix. 

\begin{algorithm}[t]
\SetAlgoLined
\caption{HeteroFedTTA}
\label{alg:hetero_fedtta}
\textbf{Input}: public unlabeled dataset $\mathcal{D}_p$, total communication rounds $R$ and local iterations $\tau$ \\
\textbf{[Training]} \\
\For{\texttt{each round} $r$ : 0 \texttt{to} $R-1$} {
    // Local Training \\
    \textbf{Distribute:} 
    Each client downloads the updated ensemble knowledge $\mathcal{F}_{base}(\mathcal{D}_p)$ and $\mathcal{F}_{per}(\mathcal{D}_p)$ \\
    \textbf{Digest:} Each client distills the ensemble knowledge through the KD loss defined in Eq.~\ref{eq:kd_loss} \\
    \textbf{Revisit:} Each client meta-trains its local models on its own labeled dataset for $\tau$ iterations \\
    \textbf{Communicate:} Each client computes the logits on the public dataset made by the base prediction model and personalized prediction model, 
    i.e., $f(\mathcal{D}_p|\psi^i)$ and $f(\mathcal{D}_p|\tilde{\psi^i})$,
    and transmits the results to the central server \\
    // Server Aggregation \\
    \textbf{Aggregate}: 
    The server updates the ensemble knowledge 
    $\mathcal{F}_{base}(\mathcal{D}_p) = \frac{1}{N} \sum_{i} f(\mathcal{D}_p|\psi^{i})$ and 
    $\mathcal{F}_{per}(\mathcal{D}_p) = \frac{1}{N} \sum_{i} f(\mathcal{D}_p|\tilde{\psi^i})$ \\ 
}
\textbf{[Test of client $c_{new}$]} \\
\textbf{Digest:} 
Client $c_{new}$ downloads the ensemble knowledge $\mathcal{F}_{base}(\mathcal{D}_p)$ and $\mathcal{F}_{per}(\mathcal{D}_p)$ and 
distills it to its local models 
through the KD loss defined in Eq.~\ref{eq:kd_loss} \\
\textbf{Revisit:} Client $c_{new}$ adapts its base prediction model with the adaptation model and makes predictions with the personalized prediction model 
\end{algorithm}

\subsection{Summary}
In Algorithm~\ref{alg:algorithm}, 
we provide a detailed description of the training and testing procedures of \algo~and \algopp. 
During training, 
\algo~is trained solely on the training clients without accessing any data from new clients. 
During testing, 
a well-trained adaptation model 
personalizes the base prediction model 
for the new clients 
based on their unlabeled datasets. 
\algopp~improves \algo's performance 
through two simple strategies: 
regularizing the outputs of the base prediction model and 
early stopping the adaptation based on entropy. 

The detailed algorithm of \heteroalgo~
is provided in Algorithm~\ref{alg:hetero_fedtta}. 
\heteroalgo~extends the application of \algo~and \algopp~to device heterogeneous settings 
by transferring the knowledge of various base prediction models and adaptation models 
through knowledge distillation on a public dataset.

\section{Experiments}
In this section, 
we first introduce the experiment setup 
and then compare \algo~and its variants 
against eleven representative baselines to date 
on five popular benchmark datasets. 
Then we perform additional experiments 
to thoroughly investigate the effectiveness of our proposed methods 
in various testing environments, 
including 
varying degrees of distribution shift, 
varying volumes of test datasets, and concept shift.
Finally, we run \heteroalgo~ with varying $\lambda$ 
to validate the effectiveness of our proposed KD loss.

\subsection{Experimental Settings}
\subsubsection{Datasets}
We conduct experiments on five benchmark datasets, 
namely MNIST~\cite{lecun1998gradient}, 
CIFAR-10~\cite{cifar10}, 
FEMNIST~\cite{caldas2018leaf}, 
CIFAR-100~\cite{cifar100}, 
and Fashion-MNIST~\cite{xiao2017fashion}. 
To partition MNIST, CIFAR-10, and Fashion-MNIST, 
we employ the pathological partitioning method 
proposed by ~\citet{mcmahan2017communication}. 
Specifically, we divide each dataset into 100 clients, where each client has at most two labels among the ten. 
For CIFAR-100, which contains twenty superclasses, each superclass has five fine-grained classes, and we evenly distribute the data of each superclass to five clients, resulting in a total of 100 clients. 
We sample 185 clients from FEMNIST, which is a naturally heterogeneous dataset with varying writing styles from client to client. 
We randomly select 50\% of the clients as training clients 
and the remaining 50\% as test clients. 
For each training client, 
we partition the local dataset 
into train and validation splits with a ratio of 85:15. 
We present detailed client statistics for all datasets in the appendix.

\subsubsection{Baselines}
{In the experiments, we consider eleven baselines}, which include 
generic FL methods 
(\method{FedAvg}~\cite{mcmahan2017communication}, 
\method{FedProx}~\cite{li2020federated}, 
and \method{SCAFFOLD}~\cite{karimireddy2020scaffold}
), 
generalized FL methods 
(\method{AFL}~\cite{mohri2019agnostic}, 
\method{FedGMA}~\cite{tenison2022gradient}, 
\method{FedADG}~\cite{zhang2021federated}, 
and \method{FedSR}~\cite{nguyen2022fedsr}), 
and the most related method to us \method{ODPFL-HN}~\cite{amosy2021inference}. 
In addition, 
we adapt 
a centralized test-time adaptation method 
\method{TENT}~\cite{wang2020tent} with appropriate adjustments 
to our investigated \setting~ setting. 
We also compare \algo~ with two PFL-based methods: \method{PFL-Sampled} and \method{PFL-Ensemble}. 
\method{PFL-Sampled} selects a personalized model randomly 
from the training clients and sends it to the new client, 
while \method{PFL-Ensemble} sends all models from the training clients to the new client, 
and predictions are made by averaging the logits of all models. 
We implement \method{PFL-Sampled} and \method{PFL-Ensemble} based on \method{FedPer}~\cite{arivazhagan2019federated}, 
a simple yet strong PFL method. 
Note that we use \method{PFL-Sampled} and \method{PFL-Ensemble} only as baselines, 
as sending training clients' models to test clients 
could compromise the privacy of the training clients 
and \method{PFL-Ensemble} brings heavy communication and computation costs. 
We provide a detailed description of all baselines in the appendix.

\subsubsection{Implementation Details}
We implement \algo~ and its variants 
using PyTorch and train them with the SGD optimizer. 
We perform 200, 1000, 200, 1500, and 300 global communication rounds 
for MNIST, CIFAR-10, FEMNIST, CIFAR-100, and Fashion-MNIST, respectively. 
For all datasets, we set the number of local iterations $\tau$ to 20 
and the local minibatch size $B$ to 64. 
{All clients participate in the training process. }
We adopt the same neural network model for all methods on the same dataset. 
Specifically, we use a fully connected network with two hidden layers of 200 units for MNIST. 
Following \cite{mcmahan2017communication}, we use a ConvNet\cite{lecun1998gradient} 
that contains two convolutional layers and two fully connected layers for the other four datasets. 
We implement the adaptation model with a lightweight fully connected network 
that has three hidden layers of 32 units for all datasets. 
All experiments are conducted three times with different random seeds, 
and we report the mean and variance of the results. 
For each run, we report 
the validation accuracy and the test accuracy, 
where the validation accuracy reflects the performance of the training clients 
and the test accuracy reflects that of the new (test) clients. 
Particularly, we report the test accuracy of the model 
\textit{with the best performance on the validation dataset}. 
All experiments are run on six Tesla V100 GPUs. 
We perform a hyperparameter search for all methods 
and provide the resulting optimal hyperparameters. 
Due to the page limit, we defer further detailed implementation details to the appendix. 

\begin{table*}[htbp!]
    \centering
    \caption{Comparison of \algo~and its variants with baselines in terms of top-1 accuracy. 
    The best and second best results are marked with {boldface} and {underline}, respectively. 
    {In \setting~ task, we mainly focus on the evaluation metric of test accuracy.}}
    \resizebox{\linewidth}{!}{
        \begin{tabular}{l c c c c c c c c c c} 
        \toprule 
        \midrule
        \multirow{2}{*}{Method} & \multicolumn{2}{c}{MNIST} & \multicolumn{2}{c}{CIFAR-10} & \multicolumn{2}{c}{FEMNIST} & \multicolumn{2}{c}{CIFAR-100} & \multicolumn{2}{c}{Fashion-MNIST} \\ 
        \cmidrule{2-11} 
        & Validation & Test & Validation & Test & Validation & Test & Validation & Test & Validation & Test \\
        \midrule   
        \method{FedAvg}     & 95.06 $\pm$ 0.07 & 94.69 $\pm$ 0.08 
                            & 66.93 $\pm$ 0.47 & 62.81 $\pm$ 0.09 
                            & 83.98 $\pm$ 0.14 & 63.62 $\pm$ 0.11 
                            & 31.10 $\pm$ 0.15 & 23.32 $\pm$ 0.23 
                            & 90.32 $\pm$ 0.52 & 86.99 $\pm$ 0.40 \\
        \method{FedProx}    & 96.27 $\pm$ 0.12 & 95.77 $\pm$ 0.07 
                            & 66.96 $\pm$ 0.37 & 62.74 $\pm$ 0.12 
                            & 83.85 $\pm$ 0.23 & 64.99 $\pm$ 0.20 
                            & 31.16 $\pm$ 0.49 & 23.37 $\pm$ 0.17 
                            & 90.35 $\pm$ 0.65 & 87.10 $\pm$ 0.55 \\
        \method{SCAFFOLD}   & 97.48 $\pm$ 0.10 & 97.26 $\pm$ 0.07 
                            & 70.35 $\pm$ 0.27 & 66.52 $\pm$ 0.01 
                            & \best{85.89 $\pm$ 0.17} & 67.10 $\pm$ 0.30 
                            & 33.31 $\pm$ 0.38 & 25.53 $\pm$ 0.02 
                            & 90.18 $\pm$ 0.16 & 87.28 $\pm$ 0.10 \\
        \midrule
        \method{AFL}        & 96.45 $\pm$ 0.15  & 96.01 $\pm$ 0.11  
                            & 65.56 $\pm$ 0.30  & 62.05 $\pm$ 0.34  
                            & 83.72 $\pm$ 0.31  & 64.91 $\pm$ 0.57  
                            & 30.65 $\pm$ 1.02  & 22.74 $\pm$ 0.16 
                            & 89.93 $\pm$ 0.36  & 87.33 $\pm$ 0.37 \\
        \method{FedGMA}     & 95.94 $\pm$ 0.10  & 95.35 $\pm$ 0.02  
                            & 65.76 $\pm$ 0.28  & 61.91 $\pm$ 0.33  
                            & 83.74 $\pm$ 0.23  & 64.80 $\pm$ 0.35  
                            & 31.03 $\pm$ 0.29  & 23.07 $\pm$ 0.12 
                            & 89.72 $\pm$ 0.28  & 86.63 $\pm$ 0.49 \\
        \method{FedADG}     & 96.27 $\pm$ 0.12  & 95.93 $\pm$ 0.02  
                            & 67.00 $\pm$ 0.59  & 63.58 $\pm$ 0.21  
                            & 84.03 $\pm$ 0.18  & 65.49 $\pm$ 0.20  
                            & 31.73 $\pm$ 0.37  & 23.62 $\pm$ 0.19 
                            & 90.75 $\pm$ 0.11  & 87.96 $\pm$ 0.14 \\
        \method{FedSR}      & 93.97 $\pm$ 0.05  & 93.08 $\pm$ 0.14  
                            & 59.93 $\pm$ 0.25  & 53.56 $\pm$ 0.10 
                            & 80.35 $\pm$ 0.25  & 64.28 $\pm$ 0.09 
                            & 28.53 $\pm$ 0.28  & 21.31 $\pm$ 0.11 
                            & 83.71 $\pm$ 0.11  & 77.91 $\pm$ 0.42 \\
        \midrule
        \method{PFL-Sampled}    & \best{99.02 $\pm$ 0.03} & \textcolor{white}{0}9.43 $\pm$ 0.77 
                                & 83.16 $\pm$ 0.34 & \textcolor{white}{0}9.02 $\pm$ 0.44 
                                & 65.34 $\pm$ 0.27 & \textcolor{white}{0}1.81 $\pm$ 0.06 
                                & \best{52.35 $\pm$ 0.04} & \textcolor{white}{0}0.76 $\pm$ 0.03 
                                & \best{97.03 $\pm$ 0.19} & 13.00 $\pm$ 2.00 \\
        \method{PFL-Ensemble}   & \best{99.02 $\pm$ 0.03} & 44.82 $\pm$ 1.03 
                                & 83.16 $\pm$ 0.34 & 30.84 $\pm$ 1.23 
                                & 65.34 $\pm$ 0.27 & 46.96 $\pm$ 1.04 
                                & \best{52.35 $\pm$ 0.04} & \textcolor{white}{0}5.94 $\pm$ 0.08 
                                & \best{97.03 $\pm$ 0.19} & 50.16 $\pm$ 3.01 \\
        \method{ODPFL-HN}       & 96.39 $\pm$ 0.33 & 95.67 $\pm$ 0.07 
                                & 68.82 $\pm$ 1.34 & 58.39 $\pm$ 1.00 
                                & 74.07 $\pm$ 0.25 & 65.47 $\pm$ 0.38
                                & 33.33 $\pm$ 0.12 & 27.20 $\pm$ 0.31
                                & 89.24 $\pm$ 0.37 & 85.27 $\pm$ 0.34 \\
        \method{TENT}           & 98.44 $\pm$ 0.05 & 98.51 $\pm$ 0.03 
                                & 77.53 $\pm$ 0.32 & 74.46 $\pm$ 0.19
                                & 76.06 $\pm$ 0.27 & 58.31 $\pm$ 0.72
                                & 40.18 $\pm$ 0.18 & 32.11 $\pm$ 0.31
                                & 95.68 $\pm$ 0.07 & \second{94.98 $\pm$ 0.33} \\
        \midrule
        \algo                   & 98.73 $\pm$ 0.10 & 98.47 $\pm$ 0.15 
                                & 82.04 $\pm$ 0.84 & 79.80 $\pm$ 0.39
                                & 84.64 $\pm$ 0.13 & \second{69.05 $\pm$ 0.40}
                                & 42.16 $\pm$ 2.07 & 35.53 $\pm$ 1.91
                                & 95.29 $\pm$ 0.70 & 93.26 $\pm$ 2.42 \\
        \algoprox               & 98.76 $\pm$ 0.18 & \second{98.64 $\pm$ 0.03} 
                                & \second{83.24 $\pm$ 0.58} & \second{81.41 $\pm$ 0.17}
                                & \second{84.85 $\pm$ 0.11} & \best{69.85 $\pm$ 0.23}
                                & 43.64 $\pm$ 1.68 & \second{36.52 $\pm$ 1.53}
                                & 95.53 $\pm$ 0.46 & 94.60 $\pm$ 0.99 \\
        \algopp                 & \second{98.94 $\pm$ 0.11} & \best{98.66 $\pm$ 0.01} 
                                & \best{85.82 $\pm$ 0.38} & \best{83.31 $\pm$ 0.27} 
                                & 83.78 $\pm$ 0.12 & 62.00 $\pm$ 0.11
                                & \second{48.59 $\pm$ 0.80} & \best{42.95 $\pm$ 0.50}
                                & \second{96.04 $\pm$ 0.17} & \best{96.48 $\pm$ 0.25} \\
        \midrule
        \bottomrule
        \end{tabular}
    }
    \label{tab:overall-performance}
\end{table*}

\begin{table*}[htbp]
\caption{
Comparison of \algo~ and its variant with baselines under different degrees of distribution shift. 
    The best and second best results are marked with {boldface} and {underline}, respectively.}
\resizebox{0.8\linewidth}{!}{
\begin{tabular}{|l|ll|l|l|l|llllll|}
\cline{1-3} \cline{5-12}
\multicolumn{1}{|c|}{\multirow{2}{*}{Method}} & \multicolumn{2}{c|}{Acc}                                    
&  & 
\multirow{2}{*}{Method} & \multirow{2}{*}{Validation} & \multicolumn{6}{c|}{Test ($\alpha$)}                                                                                                                                                                                           \\ \cline{2-3} \cline{7-12} 
\multicolumn{1}{|c|}{}  
& \multicolumn{1}{c|}{Validation} 
& \multicolumn{1}{c|}{Test} 
&   &   &       
& \multicolumn{1}{c|}{$\alpha=$ 0.01} 
& \multicolumn{1}{c|}{$\alpha=$ 0.03} 
& \multicolumn{1}{c|}{$\alpha=$ 0.05} 
& \multicolumn{1}{c|}{$\alpha=$ 0.1} 
& \multicolumn{1}{c|}{$\alpha=$ 0.3} 
& \multicolumn{1}{c|}{$\alpha=$ 0.5} 
\\ 
\cline{1-3} \cline{5-12} 
\method{FedAvg}                                        
& \multicolumn{1}{c|}{69.06} 
& \multicolumn{1}{c|}{68.34} 
&  
& \method{PFL-Sampled}             
& \multicolumn{1}{c|}{82.37}                         
& \multicolumn{1}{c|}{\textcolor{white}{0}7.14}   
& \multicolumn{1}{c|}{12.65}   
& \multicolumn{1}{c|}{10.38}   
& \multicolumn{1}{c|}{23.41}  
& \multicolumn{1}{c|}{11.00}  
& \multicolumn{1}{c|}{11.68}  
\\ \cline{1-3} \cline{5-12} 
\method{FedProx}                                       
& \multicolumn{1}{c|}{68.88} 
& \multicolumn{1}{c|}{68.35} 
& 
& \method{PFL-Ensemble}            
& \multicolumn{1}{c|}{82.50} 
& \multicolumn{1}{c|}{47.24}     
& \multicolumn{1}{c|}{48.64}     
& \multicolumn{1}{c|}{44.77}     
& \multicolumn{1}{c|}{42.93}    
& \multicolumn{1}{c|}{38.06}    
& \multicolumn{1}{c|}{38.25}    
\\ \cline{1-3} \cline{5-12} 
\method{SCAFFOLD}                                      
& \multicolumn{1}{c|}{72.01} 
& \multicolumn{1}{c|}{71.07} 
&  
& \method{TENT}                    
& \multicolumn{1}{c|}{82.82} 
& \multicolumn{1}{c|}{\second{91.48}}     
& \multicolumn{1}{c|}{\second{88.81}}     
& \multicolumn{1}{c|}{85.40}     
& \multicolumn{1}{c|}{84.23}    
& \multicolumn{1}{c|}{74.37}    
& \multicolumn{1}{c|}{71.30}    
\\ \cline{1-3} \cline{5-12} 
\method{AFL}                                           
& \multicolumn{1}{c|}{66.72} 
& \multicolumn{1}{c|}{66.21} 
&  
& \method{ODPFL-HN} 
& \multicolumn{1}{c|}{66.53} 
& \multicolumn{1}{c|}{68.06} 
& \multicolumn{1}{c|}{67.88} 
& \multicolumn{1}{c|}{67.03} 
& \multicolumn{1}{c|}{64.60} 
& \multicolumn{1}{c|}{63.32}    
& \multicolumn{1}{c|}{61.52}    
\\ 
\cline{1-3} \cline{5-12} 
\method{FedGMA}                                        
& \multicolumn{1}{c|}{60.98} 
& \multicolumn{1}{c|}{60.78} 
&  
& \algo                  
& \multicolumn{1}{c|}{83.39} 
& \multicolumn{1}{c|}{89.24}     
& \multicolumn{1}{c|}{87.15}     
& \multicolumn{1}{c|}{84.76}     
& \multicolumn{1}{c|}{83.34}    
& \multicolumn{1}{c|}{76.24}    
& \multicolumn{1}{c|}{\second{74.49}}    
\\ \cline{1-3} \cline{5-12} 
\method{FedADG} 
& \multicolumn{1}{c|}{69.04} 
& \multicolumn{1}{c|}{68.51} 
&  
& \algoprox
& \multicolumn{1}{c|}{\second{83.89}} 
& \multicolumn{1}{c|}{90.82} 
& \multicolumn{1}{c|}{88.43}     
& \multicolumn{1}{c|}{\second{85.97}}     
& \multicolumn{1}{c|}{\second{84.40}}
& \multicolumn{1}{c|}{\second{76.95}}    
& \multicolumn{1}{c|}{\best{74.93}}    
\\ \cline{1-3} \cline{5-12} 
\method{FedSR}                                         
& \multicolumn{1}{c|}{61.79} 
& \multicolumn{1}{c|}{61.98}
&  
& \algopp 
& \multicolumn{1}{c|}{\best{86.11}}
& \multicolumn{1}{c|}{\best{94.49}}     
& \multicolumn{1}{c|}{\best{90.01}}     
& \multicolumn{1}{c|}{\best{87.70}}     
& \multicolumn{1}{c|}{\best{85.86}}    
& \multicolumn{1}{c|}{\best{76.98}}    
& \multicolumn{1}{c|}{73.20}    
\\ \cline{1-3} \cline{5-12} 
\end{tabular}
}
\label{tab:distribution-shift}
\end{table*}

\subsection{Overall Performance}
We present the validation accuracy and test accuracy of all methods on all datasets in Tab.~\ref{tab:overall-performance}. 
It shows that 
\algoprox~ performs better than \algo~ on all datasets 
and \algopp~ further improves \algoprox~ by a large margin on all datasets except FEMNIST. 
These observations confirm the effectiveness of our introduced components. 
Moreover, \algopp~ beats all the competitors in terms of test accuracy across all datasets, 
exceeding the best baseline by approximately 9\%, 2\%, 10\% and 1.5\% on CIFAR-10, FEMNIST, CIFAR-100 and Fashion-MNIST, respectively. 
The {PFL} method optimizes for training clients. 
In terms of validation accuracy, 
the PFL-based methods achieve the best results on three out of five datasets, 
while \method{SCAFFOLD} performs best on FEMNIST. 
Our proposed methods outperform others on CIFAR-10 and achieves the second-best results on other datasets. 
The experiment results demonstrate that while we aim to personalize models for new clients, 
our proposed methods still perform well in training clients. 
Due to page limit,  we defer the comparison and analysis  of all methods 
in terms of learning efficiency to the appendix.

\subsection{Effects of Distribution Shift}
Distribution shift between training clients and test clients is challenging for \setting. 
To verify the effectiveness of our proposed methods under different degrees of distribution shift, 
following the previous study~\cite{hsu2019measuring}, 
we generate training and test clients with different distribution shift using Dirichlet distribution with different concentration parameters $\alpha$. 
Specifically, we partition the train set of CIFAR-10 into 50 training clients with a $\alpha$ of 0.1 
and the test set of CIFAR-10 into 50 test clients with $\alpha \in [0.01, 0.03, 0.05, 0.1, 0.3, 0.5]$. 

The experiment results are shown in Tab~\ref{tab:distribution-shift}. 
\method{FedAvg}, \method{FedProx}, \method{SCAFFOLD}, and the \method{GFL} methods 
output the same model for all clients, 
and the changes in the test environment will not affect the test performance. 
For \method{PFL-Sampled}, when the test environment is consistent with the training environment ($\alpha=0.1$), it obtains its highest test accuracy. 
When the $\alpha$ is smaller, 
the characteristics of client data distribution is more obvious, 
which is more beneficial to \method{TENT} and \algo~and its variants. 
Overall, our proposed methods outperform existing baselines under different degrees of distribution shift.

\subsection{Effects of Dataset Size}
In \algo, 
the adaptation model generates personalized models for clients 
by leveraging information from their unlabeled data. 
However, 
in federated learning, 
the new clients often have limited data points, 
which restricts the amount of knowledge 
they can convey about the underlying data distribution. 
This poses significant challenges for \algo~and the baselines (\method{ODPFL-HN} and \method{TENT}). 
To demonstrate the robustness of our approach against sample sizes, 
we create more challenging test environments by reducing the data volume available to the test (new) clients. 

\begin{figure}[t!]
    \centering
    \includegraphics[width=1.0\linewidth]{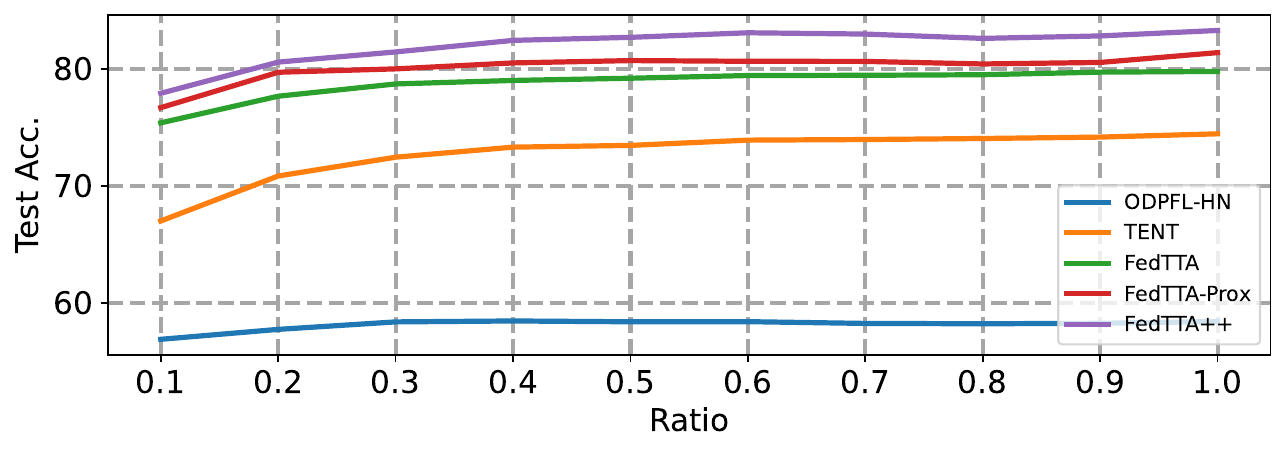}
    \caption{Each curve depicts the relationship between the size of test client datasets and the average test accuracy. 
    In general, the larger the client dataset, the higher the test accuracy.}
    \label{fig:size-of-test}
\end{figure}

Specifically, we evaluate the performance of \algo, \algoprox, \algopp, \method{TENT}, and \method{ODPFL-HN} on CIFAR-10 dataset. 
For each test client, we reduce the amount of data to $\alpha$ times its original amount, 
where $\alpha$ ranges from 0.1 to 1.0. 
The results are presented in Fig.~\ref{fig:size-of-test}, 
which shows the average accuracy achieved by the test clients as $\alpha$ varies. 
It can be observed that our proposed methods outperform the competitors across all settings. 
Notably, both \algo~, \algoprox, and \algopp~demonstrate high effectiveness in data-scarce scenarios. 
Even when the data volume is reduced to 0.1 times the original (60 samples per client), \algo~, \algoprox~and \algopp~only experience slight performance drops, from 79.80, 81.41 and 83.31 to 75.40, 76.70 and 77.93 respectively, which is still significantly higher than all baselines. 

\subsection{Concept Shift}
\textit{Concept shift} is also prevalent in federated learning, 
where the conditional distribution $p(x|y)$ varies among clients. 
We conduct additional experiments to demonstrate the effectiveness of \algo~in the presence of concept shift during testing. 
Specifically, following \method{FedSR}~\cite{nguyen2022fedsr}, 
we create five domains with concept-shift among them: 
$M_{0}$, $M_{15}$, $M_{30}$, $M_{45}$, and $M_{60}$ 
by rotating the MNIST~\cite{lecun1998gradient} dataset 
counterclockwise at angles of [0, 15, 30, 45, 60] degrees. 
We set up 50 training clients with $M_{0}$, $M_{30}$, and $M_{60}$ while 50 test clients with $M_{15}$ and $M_{45}$. 
Please refer to the appendix for detailed statistics of the clients' datasets. 

\begin{figure}[b!]
    \centering
    \includegraphics[width=1.0\linewidth]{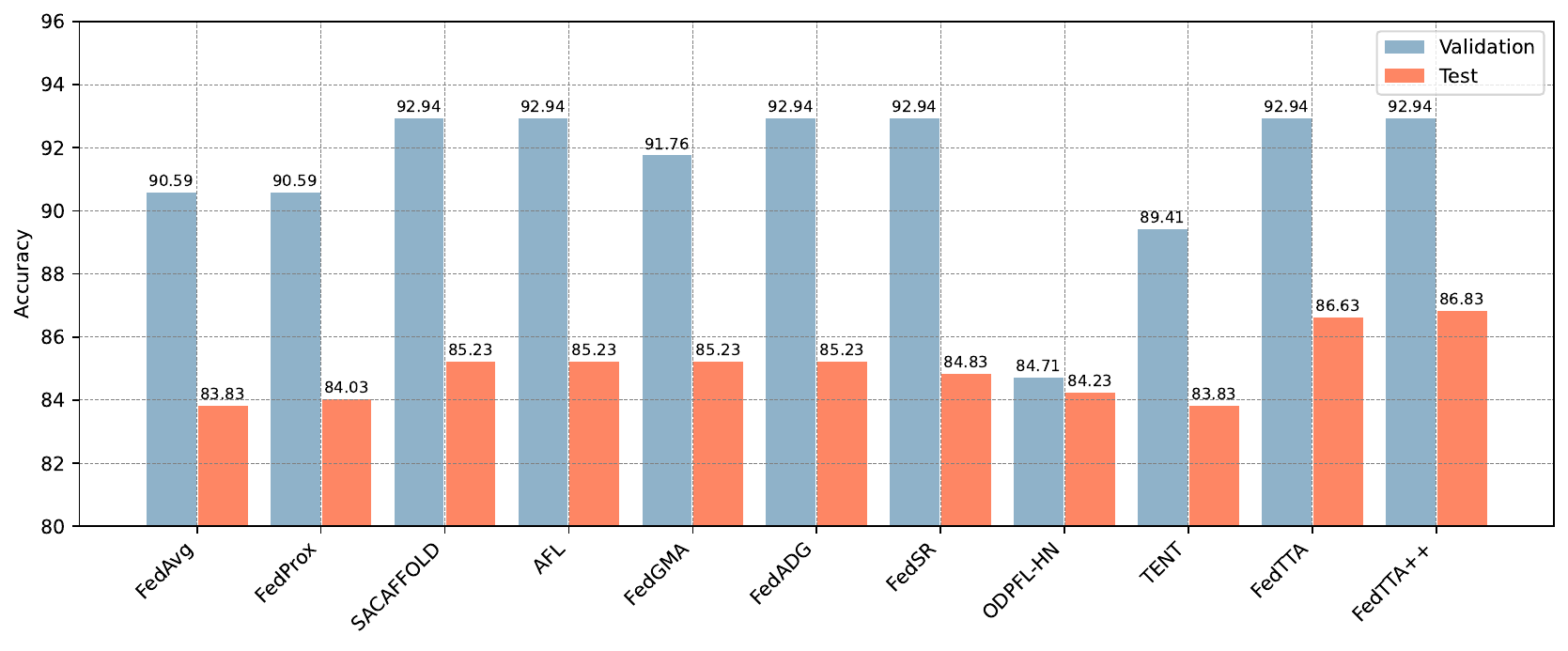}
    \caption{Performance comparison in the presence of concept shift. 
    For better illustration, we exclude the results of \method{PFL-Sampled} and \method{PFL-Ensemble} here, 
    as both methods exhibit very poor performance.}
    \label{fig:concept-shift}
\end{figure}

We present the experiment results in Fig.~\ref{fig:concept-shift}. 
Significant concept shift 
makes it challenging to construct a well-perform ensemble model 
for test clients 
from 
the training clients. 
\method{PFL-Sampled} achieves a validation accuracy of 40.00 and a test accuracy of 9.92, 
while \method{PFL-Ensemble} achieves a validation accuracy of 67.06 and a test accuracy of 63.47. 
For better illustration, we exclude the results of \method{PFL-Sampled} and \method{PFL-Ensemble} from Fig.~\ref{fig:concept-shift}, 
as both methods exhibit very poor performance.

Fig.~\ref{fig:concept-shift} clearly demonstrates the superior performance of our proposed methods. 
Specifically, on the test datasets, \algo~ and \algopp~ outperform the best baselines by 1.4\% and 1.6\%, respectively. 
Additionally, on the validation datasets, \algo~ and \algopp~ achieve results that are comparable to the best baselines. 

\subsection{Heterogeneous Settings} 
In this subsection, 
we will investigate the effectiveness of our proposed KD loss function (Eq.~\ref{eq:kd_loss}) 
by running \heteroalgo~on CIFAR-10 dataset with varying hyperparameter $\lambda \in \{0.0, 0.2, 0.4, 0.6, \\ 0.8, 1.0\}$. 
It is worth noting that when $\lambda$ equals zero, the loss degrades to the vanilla KD loss for \algo. 
For simplicity, 
we implement \heteroalgo~ based on \algo~ 
and use the same data partitioning 
as in the main experiment. 
We generate three types of models 
with different complexity levels 
by controlling the hidden channels, 
and we assign each model to a client 
with equal probability. 
$\eta_{inner}$, $\eta_{outer}$ and $\eta_{adapt}$ are set to 0.5, 0.1 and 0.005, respectively. 
For the detailed model structure, please refer to the appendix. 
\method{TENT} achieves the second best result after \algo~ on CIFAR10 (see Table~\ref{tab:overall-performance}). 
We additionally implement heterogeneous \method{TENT} as a strong baseline by directly integrating \method{TENT} into the \method{FedMD} framework. 
We do not compare with \method{ODPFL-HN} in this experiment since the hypernetwork produces the same model structure for all clients. 

\begin{figure}[t!]
    \centering
    \includegraphics[width=1.0\linewidth]{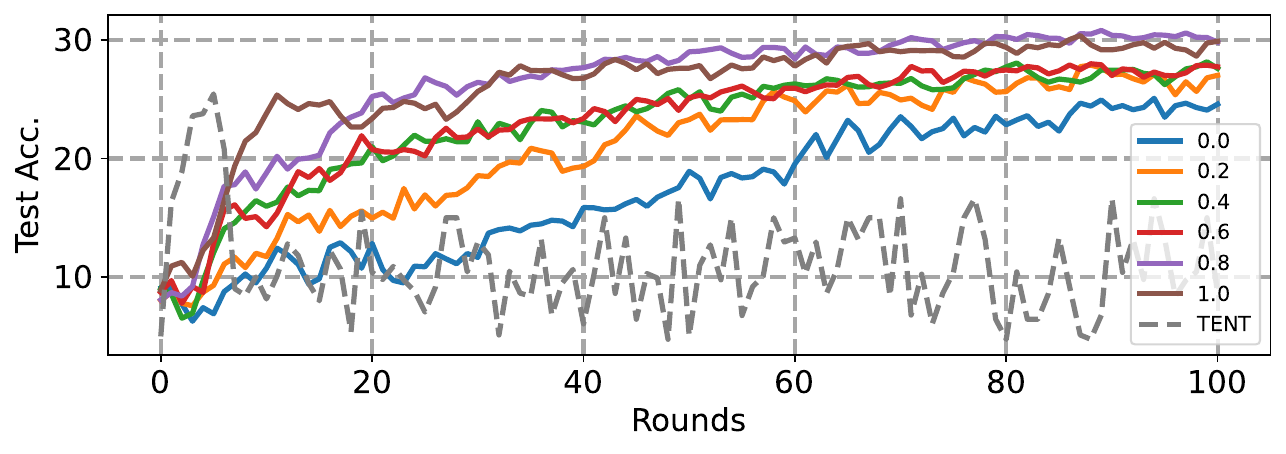}
    \caption{Test accuracy curves of \method{TENT} and \heteroalgo~ with different $\lambda$ values.}
    \label{fig:kd-acc-curves}
\end{figure}

Fig.~\ref{fig:kd-acc-curves} 
shows the test accuracy curves 
of \method{TENT} and \heteroalgo~\\with different $\lambda$. 
The results indicate that 
\heteroalgo~ is more effective and stable than \method{TENT}. 
Moreover, increasing the beta value generally improves the distillation effect. 
When $\lambda$ is approximately 0.8, it achieves the best performance, 
surpassing the vanilla KD by a significant margin. 
Overall, the experiment result validates the effectiveness of our proposed KD loss.

\section{Conclusion}
\label{sec:con}
In this paper, we investigated a practical yet challenging task for heterogeneous federated learning, i.e., Unsupervised Personalized Federated Learning towards new clients (UPFL), which aims to provide personalized models for unlabeled new clients after the federated model has been trained and deployed. To address the task, we first proposed a base method \algo, and we then improved \algo~ with two simple yet effective optimization strategies: 
enhancing the adaptation model with regularization on the base prediction model 
and early-stopping the adaptation process through entropy. 
Last, we suggested a novel knowledge distillation loss for the special architecture of \algo~and heterogeneous model setting. 
We conducted extensive experiments on five datasets and compare \algo~and its variants against eleven baselines. 
The experimental results demonstrate the effectiveness of our proposed methods in addressing the \setting~task. 
\clearpage
\bibliographystyle{ACM-Reference-Format}
\bibliography{sample-base}

\clearpage
\appendix

\setcounter{section}{0}
\setcounter{figure}{0}
\setcounter{table}{0}

\section{Knowledge Distillation}
\begin{figure}[htbp!]
    \centering
    \includegraphics[width=0.5\textwidth,trim=135 180 350 40,clip]{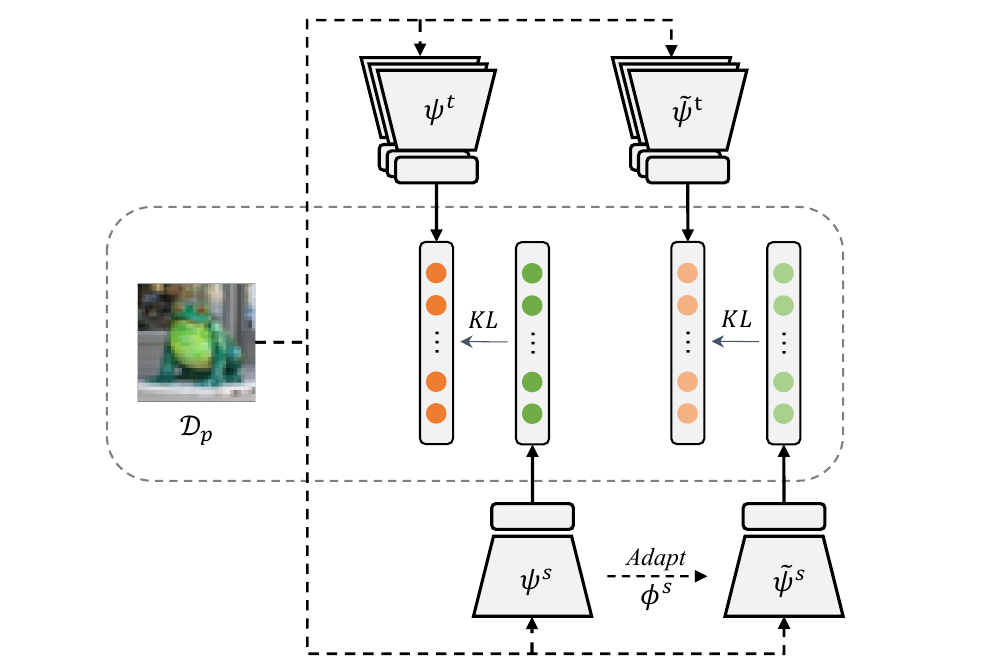}
    \caption{
    Schematic diagram of our proposed knowledge distillation loss in \heteroalgo. $\mathcal{D}_p$ is the public unlabeled dataset.
    }
    \label{fig:kd-framework}
\end{figure}

We provide a schematic diagram of our proposed knowledge distillation loss in \heteroalgo~in Fig.\ref{fig:kd-framework}. 
$\mathcal{D}_p$ is the public unlabeled dataset. 
$\{\psi^{s}, \tilde{\psi}^{s}\}$ and $\{\psi^{t}, \tilde{\psi}^{t}\}$ are the parameters of the base prediction model and the personalized prediction model of the student and the teacher, respectively, and $\phi^{s}$ is the parameters of the adaptation model of the student. 

During the knowledge distillation process, the student should 
enforce its base prediction model and adaptation model learn from the corresponding models of the teacher, respectively. 
Specifically, 
$\psi^{s}$ should mimic the outputs of $\psi^{t}$, 
while $\phi^{s}$ should learn the personalization capability of $\phi^{t}$. 
Under the constraint of 
minimizing the KL-divergence between the outputs of $\psi^{s}$ and $\psi^{t}$, 
$\phi^{s}$ can learn the personalization ability of $\phi^{t}$ 
by minimizing the KL-divergence between the outputs of $\tilde{\psi}^{s}$ and $\tilde{\psi}^{t}$. 

\section{Datasets}
We first give a introduction to the datasets used, followed by details of the client partitions for all datasets. 
We conduct experiments on following five benchmark datasets, namely MNIST, CIFAR-10, FEMNIST, CIFAR-100 and Fashion-MNIST. 
\begin{itemize}
    \item MNIST: is a widely used handwritten digit recognition dataset, 
    which consists of 10 classes of numerical digits 0-9. 
    \item CIFAR-10: is a classic computer vision dataset that consists of 60,000 32x32 color images in 10 classes, where 50,000 images are used for training and the remaining 10,000 images are used for testing. 
    \item FEMNIST: is a naturally heterogeneous dataset, which consists of 805,263 handwritten images from 3,550 users, where each user has a unique writing style. 
    \item CIFAR-100: is a computer vision dataset that contains 60000 32x32 color images in 100 classes, where 50000 images are used for training and the remaining 10000 images are used for testing. 
    The 100 classes in this dataset are grouped into 20 superclasses. 
    Each image in the CIFAR-100 dataset is labeled with one of the 100 fine-grained classes and one of the 20 superclasses. 
    \item Fashion-MNIST: is a 10-classes computer vision dataset, which consists of 70,000 28x28 grayscale images, each of which represents a fashion item. 
\end{itemize}
For MNIST, CIFAR-10, CIFAR-100 and Fashion-MNIST, 
we union the training set and the test set, and then divide the entire data set into 100 clients in a non-IID manner. 
For MNIST, CIFAR-10 and Fahsion-MNIST, we partition the dataset by the pathological partition strategy, where each client contains at most two classes. 
For CIFAR-100, we divide the data of each superclass evenly to 5 clients, each client contains 5 classes out of 100 fine-grained classes. 
For FEMNIST, we sample 185 users from the 3,550 users for experiments. 
Then we randomly select 50\% of the clients as training clients 
and the remaining 50\% as test clients. 
For each training client, 
we partition the local dataset 
into train and validation splits with a ratio of 85:15. 
We present the details of the client partitions for all datasets in Tab.~\ref{tab:client-partitions}. 
\begin{table}[h!]
\centering
\caption{Statistics of datasets used for experiments. FEMNIST is heterogeneous by nature, while the others follow pathological non-IID partitions.
}
\resizebox{\linewidth}{!}{
\begin{tabular}{ l c c c c c c} 
\toprule
\midrule
Dataset         & Classes   & \#Clients     & \#Samples     & \#Samples per client \\
\midrule
MNIST           & 10        &  100      & 70,000       & 700 \\
CIFAR-10        & 10        &  100      & 60,000       & 600 \\
FEMNIST         & 62        &  185      & 40,272       & 218 \\
CIFAR-100       & 100       &  100      & 60,000       & 600 \\
Fashion-MNIST   & 10        &  100      & 70,000       & 700 \\
\midrule
\bottomrule
\end{tabular}
}
\label{tab:client-partitions}
\end{table}

\section{Baselines}
We compare our proposed methods against following eleven baselines. 
\begin{itemize}
    \item \method{FedAvg}: is the first-proposed FL method, which proceeds between clients' empirical risk minimization and server's model aggregation.

    \item \method{FedProx}: is a reparameterization of \method{FedAvg}, 
    which adds a proximal term to clients' local objective functions 
    to prevent significant divergence between the global model and local model. 
    
    \item \method{SCAFFOLD}: leverages control variates to accelerate the model convergence.     

    \item \method{AFL}: optimizes a centralized model for any possible target distribution 
    formed by a mixture of the client distributions. 
    
    \item \method{FedGMA}: proposes a gradient-masked averaging approach for federated learning 
    that promotes the model to learn invariant mechanisms across clients. 
    
    \item \method{FedADG}: employs the federated adversarial learning approach to measure and align the distributions among different source domains
via matching each distribution to a adaptively generated reference distribution. 

    \item \method{FedSR}: enforces 
    an L2-norm regularizer on the representation 
    and a conditional mutual information regularizer 
    to encourage the model to only learn essential information. 

    \item \method{PFL-Sampled}: selects a personalized model randomly from the training clients and sends it to the new client. 
    
    \item \method{PFL-Ensemble}: sends all models from the training clients to the new client, and predictions are made by averaging the logits of all models. 
    \item \method{ODPFL-HN}: simultaneously learns a client encoder network and a hyper-network, 
    where the client encoder network takes the client's unlabelled data as input 
    and outputs the client representation, 
    and the hyper-network generates personalized model weights 
    based on the client's representation. 

    \item \method{TENT}: fine-tunes the model 
    by minimizing the entropy of its predictions 
    on the new client's unlabelled data. 
\end{itemize}

\begin{table}[h!]
\centering
\caption{Statistics of the clients' datasets in the concept shift environment.}
\resizebox{\linewidth}{!}{
\begin{tabular}{ l c c c c} 
\toprule
\midrule
Clients             & \#Clients     & Angles            & \#Total Samples     & \#Samples per client \\
\midrule
Training            &  25           & [0, 30, 60]       & 500                 & 20 \\
Testing             &  25           & [15, 45]            & 500               & 20 \\
\midrule
\bottomrule
\end{tabular}
}
\label{tab:concept-shift-partitions}
\end{table}

\begin{table}[htbp!]
\centering
\caption{Three types of models with different complexity levels, i.e., small, medium and big.}
\resizebox{\linewidth}{!}{
\begin{tabular}{ l c c c c} 
\toprule
\midrule
Complexity          & Prediction Model      & Adaptation Model \\
\midrule
Small               & Conv(16)-Conv(32)->MLP(512)->MLP(10)
                    & MLP(32)->MLP(32)->MLP(1) \\
Medium              & Conv(32)->Conv(64)->MLP(512)->MLP(10) 
                    & MLP(64)->MLP(64)->MLP(1) \\
Big                 & Conv(64)->Conv(128)->MLP(512)->MLP(10)
                    & MLP(128)->MLP(128)->MLP(1) \\
\midrule
\bottomrule
\end{tabular}
}
\label{tab:heterogeneous-models}
\end{table}

\section{Implementation Details}
We provide a detailed implementation details of all methods. Particularly, we perform a grid search of hyperparameters for all methods, 
and the search space for each hyperparameter for each method is as follows: 
\begin{itemize}
    \item \method{FedAvg}: 
    We search for the optimal learning rate of local optimization $\eta \in \{0.001, 0.003, 0.005, 0.01, 0.03, 0.05, 0.1, 0.3, 0.5 \}$. 
    It is important to note that unless explicitly mentioned, this search space for learning rates is applied to other methods as well.
    \item \method{FedProx}: We search for the optimal learning rate $\eta$ and coefficient of the proximal term $\mu \in \{0.001, 0.01, 0.1, 1\}$. 
    \item \method{SCAFFOLD}: We search for the optimal local learning rate $\eta_{l}$ and global learning rate $\eta_{g}$. 
    \item \method{AFL}: We search for the optimal learning rates of model parameters $\gamma_{\omega}$ and mixture coefficient $\gamma_{\lambda}$. 
    \item \method{FedGMA}: 
    We search for the optimal learning rates of local learning rate $\eta_l$ and global learning rate $\eta_g$. 
    The search space for $\eta_{g}$ is $\{0.01, 0.1, 1, 1.5, 2\}$. 
    According to the paper~\cite{tenison2022gradient}, we search for the optimal agreement threshold $\tau \in \{0.1, 0.2, 0.3, \\ 0.4, 0.5, 0.6, 0.7, 0.8, 0.9\}$. 
    \item \method{FedADG}: 
    We set the dimension of the random noise to 10. The dimension of the learned feature representation is set to 128 for MNIST and 512 for other datasets. Additionally, we search for the optimal learning rate $\eta_{f}$ for the feature extractor and the classifier.
    \item \method{FedSR}: 
    We search for the optimal learning rate $\eta$, and set the coefficients of the marginal distribution regularizer $\alpha^{L2R}$, and the conditional distribution regularizer $\alpha^{CMI}$ to their default values of 0.01 and 0.0005, respectively. 
    \item \method{PFL-Sampled} and \method{PFL-Ensemble}: We search for the optimal learning rate $\eta$. 
    \item \method{ODPFL-HN}: 
    We search for the optimal learning rates for different components: the client encoder, denoted as $\eta_{encoder}$, the hyper-network, denoted as $\eta_{hn}$, and the local optimization, denoted as $\eta_{local}$. The client embedding size is set to 32 across all datasets. 
    \item \method{TENT}: We apply the searched optimal learning rates from \method{FedAvg} to \method{TENT}. 
    \item \algo: 
    We fine-tune the inner learning rate of the prediction model, denoted as $\eta_{inner}$, the outer learning rate of the prediction model, denoted as $\eta_{outer}$, and the learning rate of the adaptation model, denoted as $\eta_{adapt}$. 
    \item \algopp: 
    In addition to searching for the optimal values of $\eta_{inner}$, $\eta_{outer}$ and $\eta_{adapt}$, 
    we also search for the optimal coefficient of the proximal term $\mu \in \{0.001, 0.01, 0.1, 1\}$. In \algopp, we propose to stop the adaptation process when the entropy of the personalized prediction model's predictions does not decrease within a certain patience value. We search for the best value of the patience from $\{1, 3, 5\}$. 
\end{itemize} 
We present the searched optimal hyperparameters in Tab.\ref{tab:hyper-parameter-details}. 
\begin{table*}[htbp!]
    \centering
    \caption{Hyper-parameter setting details of our proposed methods and the baselines on all datasets.}
    \resizebox{\linewidth}{!}{
        \begin{tabular}{|l | c | c | c | c | c |} 
        \hline
        Method              & MNIST                         
                            & CIFAR-10                  
                            & FEMNIST                           
                            & CIFAR-100 
                            & Fashion-MNIST \\
        \hline   
        \method{FedAvg}     & $\eta = 0.1$                  
                            & $\eta = 0.1$
                            & $\eta = 0.1$
                            & $\eta = 0.1$
                            & $\eta = 0.3$ \\
        \method{FedProx}    & $\eta = 0.3, \mu = 0.001$     
                            & $\eta = 0.1, \mu = 0.001$ 
                            & $\eta = 0.3, \mu = 0.001$         
                            & $\eta = 0.1, \mu = 0.001$ 
                            & $\eta = 0.3, \mu = 0.001$ \\
        \method{SCAFFOLD}   & $\eta_{l}=0.1, \eta_{g}=1.5$  
                            & $\eta_{l}=0.1, \eta_{g}=1.5$ 
                            & $\eta_{l}=0.1, \eta_{g}=1.5$ 
                            & $\eta_{l}=0.1, \eta_{g}=1.5$ 
                            & $\eta_{l}=0.1, \eta_{g}=1.5$ \\
        \hline
        \method{AFL}        & $\gamma_{\omega}=0.3, \gamma_{\lambda}=0.001$ 
                            & $\gamma_{\omega}=0.1, \gamma_{\lambda}=0.3$   
                            & $\gamma_{\omega}=0.1, \gamma_{\lambda}=0.5$ 
                            & $\gamma_{\omega}=0.1, \gamma_{\lambda}=0.3$ 
                            & $\gamma_{\omega}=0.3, \gamma_{\lambda}=0.001$ \\
        \method{FedGMA}     & $\eta_{c}=0.3, \eta_{g}=1, \tau=0.1$ 
                            & $\eta_{c}=0.1, \eta_{g}=1.5, \tau=0.1$ 
                            & $\eta_{c}=0.3, \eta_{g}=1, \tau=0.1$ 
                            & $\eta_{c}=0.1, \eta_{g}=1.5, \tau=0.1$ 
                            & $\eta_{c}=0.3, \eta_{g}=1.5, \tau=0.1$ \\
        \method{FedADG}     & $\eta_f=0.3$                    
                            & $\eta_f=0.1$                
                            & $\eta_f=0.3$ 
                            & $\eta_f=0.1$ 
                            & $\eta_f=0.3$ \\
        \method{FedSR}      & $\eta=0.3$ 
                            & $\eta=0.05$ 
                            & $\eta=0.3$ 
                            & $\eta=0.1$ 
                            & $\eta=0.3$ \\
        \hline
        \method{PFL-Sampled}& $\eta=0.1$ 
                            & $\eta=0.1$ 
                            & $\eta=0.1$ 
                            & $\eta=0.1$ 
                            & $\eta=0.3$ \\
        \method{PFL-Ensemble}   & $\eta=0.1$ 
                                & $\eta=0.1$ 
                                & $\eta=0.1$ 
                                & $\eta=0.1$ 
                                & $\eta=0.3$ \\
        \method{ODPFL-HN}   & $\eta_{local}=0.3, \eta_{encoder}=0.5, \eta_{hn}=0.5$ 
                            & $\eta_{local}=0.05, \eta_{encoder}=0.1, \eta_{hn}=0.5$
                            & $\eta_{local}=0.01, \eta_{encoder}=0.1, \eta_{hn}=0.5$ 
                            & xx
                            & $\eta_{local}=0.1, \eta_{encoder}=0.5, \eta_{hnet}=0.5$ \\
        \method{TENT}       & $\eta = 0.1$                  
                            & $\eta = 0.1$
                            & $\eta = 0.1$
                            & $\eta = 0.1$
                            & $\eta = 0.3$ \\
        \hline
        \algo               & $\eta_{inner}=0.5, \eta_{outer}=0.3, \eta_{adapt}=0.01$ 
                            & $\eta_{inner}=0.5, \eta_{outer}=0.05, \eta_{adapt}=0.005$ 
                            & $\eta_{inner}=0.3, \eta_{outer}=0.1, \eta_{adapt}=0.003$ 
                            & $\eta_{inner}=0.5, \eta_{outer}=0.1, \eta_{adapt}=0.001$
                            & $\eta_{inner}=0.05, \eta_{outer}=0.1, \eta_{adapt}=0.001$ \\
        \algoprox           & $\eta_{inner}=0.5, \eta_{outer}=0.3, \eta_{adapt}=0.001, \mu=0.01$ 
                            & $\eta_{inner}=0.3, \eta_{outer}=0.1, \eta_{adapt}=0.003, \mu=0.001$ 
                            & $\eta_{inner}=0.3, \eta_{outer}=0.1, \eta_{adapt}=0.003, \mu=0.001$ 
                            & $\eta_{inner}=0.5, \eta_{outer}=0.1, \eta_{adapt}=0.001, \mu=0.01$
                            & $\eta_{inner}=0.05, \eta_{outer}=0.1, \eta_{adapt}=0.001, \mu=0.001$ \\
        \algopp             & $patience=1$
                            & $patience=1$
                            & $patience=5$
                            & $patience=5$
                            & $patience=5$ \\
        \hline
        \end{tabular}
    }
    \label{tab:hyper-parameter-details}
\end{table*}

\section{Learning Efficiency} 
\begin{figure*}[h]
    \centering
    \includegraphics[width=1.0\textwidth]{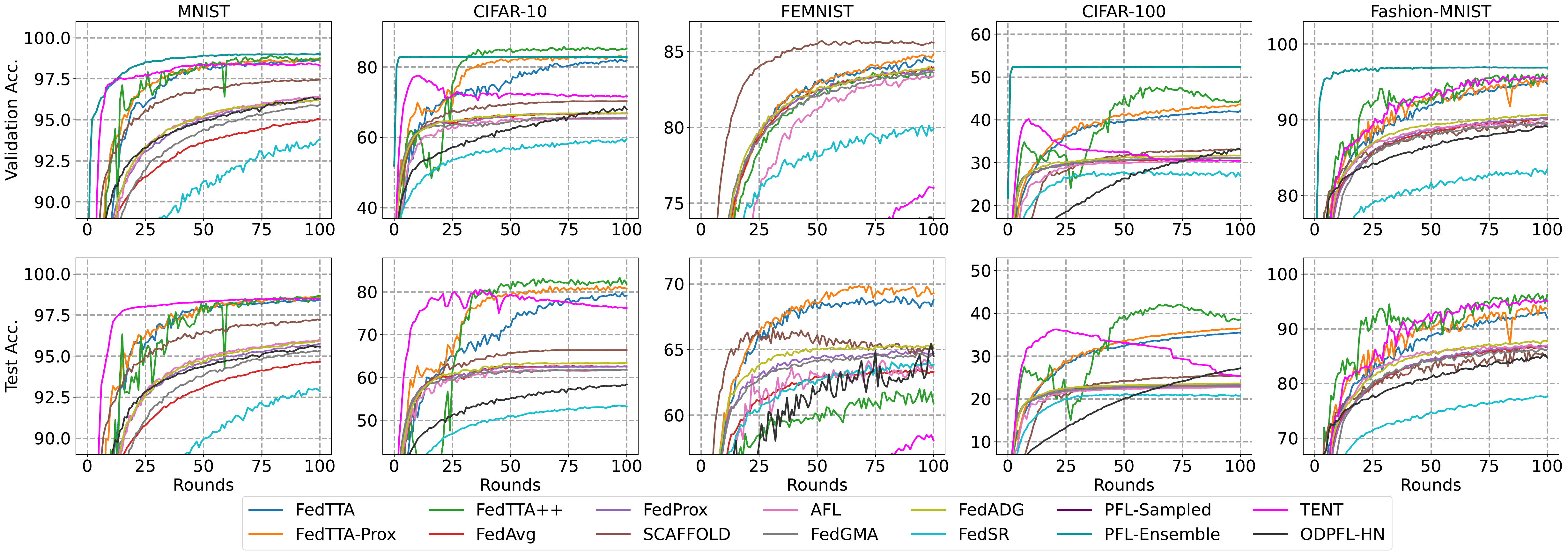}
    \caption{Convergence comparison of \algo~and its variants with the baselines. 
    Each learning curve is averaged over three random seeds.}
    \label{fig:acc_curves}
\end{figure*}

We present the accuracy curves of all methods on all datasets in Fig.~\ref{fig:acc_curves}. 
Compared with \method{FedAvg}, \method{FedProx}, and the \method{GFL} methods, 
\method{SCAFFOLD} generally converges faster and achieves higher accuracy. 
Due to distribution shift between training clients and test clients, 
\method{TENT} shows inconsistent convergence behavior on the validation set and the test set for CIFAR-10 and CIFAR-100, 
resulting in inconsistent best models on the validation and test sets. 
\algo~and \algoprox~exhibit similar convergence behavior. 
\algopp~quickly converges to the best result but is somewhat unstable in the first several communication rounds. 
We attribute this to the global base prediction model and adaptation model being undertrained 
and entropy minimization leading to overconfidence towards incorrect predictions. 
Additionally, \algo~ and its variants behave similarly on the validation and test sets. 

\section{Concept Shift} 
We randomly sample 1,000 instances from the union of the training set and test set of MNIST and allocate these samples to 50 clients in an IID (independent and identically distributed) manner. 
Among these clients, 25 are selected as training clients, and the remaining 25 are designated as testing clients. 
For the training clients, each client applies a fixed rotation angle to its local dataset, randomly selected from the set of [0, 30, 60] degrees with equal probability. 
On the other hand, the testing clients rotate their respective local datasets at a fixed angle, randomly selected from the set of [15, 45] degrees with equal probability. 
We present the partitioning information of the clients in Tab.~\ref{tab:concept-shift-partitions}. 

\section{Heterogeneous Models}
We generate three types of models 
with different complexity levels (\textit{small}, \textit{medium} and \textit{big})
by controlling the hidden channels, 
and we assign each model to a client 
with equal probability. 
We present the detailed model structure of each complexity level in Tab.~\ref{tab:heterogeneous-models}.

\end{document}